\documentclass[11pt]{article}
\usepackage{arxiv}
\usepackage[utf8]{inputenc}
\usepackage[T1]{fontenc}
\usepackage{lmodern}
\usepackage{amsmath, amssymb, amsfonts, amsthm, mathrsfs}
\usepackage{graphicx}
\graphicspath{{images/}}
\usepackage{booktabs}
\usepackage{multirow}
\usepackage{algorithm}
\usepackage{algpseudocode}
\usepackage{listings}
\usepackage[title]{appendix}
\usepackage{hyperref}
\usepackage{url}
\usepackage{doi}
\usepackage{microtype}
\usepackage{xcolor}
\usepackage[numbers]{natbib}
\date{}
\renewcommand{\today}{}

\title{Lost in translation: using global fact-checks to measure multilingual misinformation prevalence, spread, and evolution}

\author{
  Dorian Quelle\thanks{Corresponding author: \texttt{dorian.quelle@uzh.ch}} \\
  Department of Mathematical Modeling and Machine Learning, University of Zurich, Switzerland \\
  Digital Society Initiative, University of Zurich, Switzerland \\
  Oxford Internet Institute, University of Oxford, United Kingdom
  \And
  Calvin Yixiang Cheng \\
  Oxford Internet Institute, University of Oxford, United Kingdom
  \And
  Alexandre Bovet \\
  Department of Mathematical Modeling and Machine Learning, University of Zurich, Switzerland \\
  Digital Society Initiative, University of Zurich, Switzerland
  \And
  Scott A. Hale \\
  Oxford Internet Institute, University of Oxford, United Kingdom \\
  Meedan, San Francisco, United States
}

\begin{document}
\maketitle

\begin{abstract}
Misinformation and disinformation are growing threats in the digital age, affecting people across languages and borders. However, no research has investigated the prevalence of multilingual misinformation and quantified the extent to which misinformation diffuses across languages. This paper investigates the prevalence and dynamics of multilingual misinformation through an analysis of 264,487 fact-checks spanning 95 languages. To study the evolution of claims over time and mutations across languages, we represent fact-checks with multilingual sentence embeddings and build a graph where semantically similar claims are linked. We provide quantitative evidence of repeated fact-checking efforts and establish that claims diffuse across languages. Specifically, we find that while the majority of misinformation claims are only fact-checked once, 10.26\%, corresponding to more than 27,000 claims, are checked multiple times. Using fact-checks as a proxy for the spread of misinformation, we find 32.26\% of repeated claims cross linguistic boundaries, suggesting that some misinformation permeates language barriers. However, spreading patterns exhibit strong assortativity, with misinformation more likely to spread within the same language or language family. Next we show that fact-checkers take more time to fact-check claims that have crossed language barriers and model the temporal and cross-lingual evolution of claims. We analyze connected components and shortest paths connecting different versions of a claim finding that claims gradually drift over time and undergo greater alteration when traversing languages. Misinformation changes over time, reducing the effectiveness of static claim matching algorithms. The findings advocate for expanded information sharing between fact-checkers globally while underscoring the importance of localized verification.
\end{abstract}


\section{Introduction}\label{sec1}
Misinformation is a global challenge responded to in myriad local ways. The International Fact-Checking Network (IFCN) currently has 155 verified active member organizations. There have been experiments with collaborative fact-checking across countries such as \#CoronavirusFacts led by the IFCN and \#UkraineFacts led by Spanish fact-checker Maldita.es as well as collaborations within countries (e.g., \#FactsFirstPH, EKTA, and Confirma 2022). To date, however, there is no centralized repository of global fact-checks as there is for child abuse imagery (the Internet Watch Foundation) or extremist content (the Global Internet Forum to Counter Terrorism).

On the one hand, greater collaboration between fact-checking organizations could help meet the increasing demand for fact-checks. Increasing use of social media---and soon generative AI---has resulted in the volume of misinformation far exceeding the capacity of human fact-checkers \cite{konstantinovskiy2021toward, nakov2021automated, schifferes2017identifying}. Tools such as ClaimBuster, originally a model to detect check-worthy claims, aim to decrease the time that a fact-checker needs to identify and fact-check a claim \cite{adair2017progress}. Furthermore, fact-checking capacity is unequally distributed with more fact-checkers working in English than in other languages. If a large proportion of misinformation is shared across languages, centralization of fact-checks and cross-language claim matching \cite[e.g.,][]{kazemiClaim2021} could help identify misinformation even in less-resourced languages. 

On the other hand, it's not clear how often the same misinformation is spread across languages or countries. For example, there are large differences in general content across language editions of Wikipedia \cite{hale2014multilinguals, lemmerich2019world} and only a small---but very active---percentage of users author public content in multiple languages online \cite{hale2014multilinguals,haleGlobal2014}.

In this article, we investigate the extent to which misinformation claims are fact-checked by multiple fact-checking organizations (RQ1) as well as how often similar misinformation is fact-checked across different languages (RQ2). While answering these questions, we also examine the differences between content fact-checked by one vs. multiple organizations and in one vs. multiple languages.  Relatedly, we ask what languages share the most misinformation (RQ3). Next, we will investigate whether fact-checkers take longer to publish fact-checks for misinformation claims that appear in multiple languages compared to claims that only appear in a single language (RQ4). Finally, we analyze how much misinformation claims change over time (RQ5).

This paper presents an investigation into the prevalence and dynamics of multilingual misinformation through analysis of over 250,000 fact-checks in 95 languages. We find that 10.26\% of claims, corresponding to more than 27,000 fact-checks in our dataset, are checked multiple times, highlighting opportunities for greater collaboration between fact-checkers. A third of repeatedly checked claims are found in multiple languages, establishing diffusion across languages, but there is strong language assortativity. Our analysis reveals a gradual drift in claims over time with greater changes when a claim appears in multiple  languages. 

\section{Related work}\label{sec2}

\subsection{Fact-checking efforts}\label{sec2-Factchecking}
Misinformation predates the World Wide Web \cite{altay2023misinformation}, and probably existed throughout human history \cite{burkhardt2017history}. Nonetheless, misinformation and disinformation appear in scholarship about the World Wide Web in 1995, only two years after the release of the first graphical web browser: \citet{hernon1995disinformation} defines disinformation as a `deliberate attempt to deceive or mislead' and misinformation as `an honest mistake' (p. 134). As intention is difficult to reliably infer, we follow recent scholars in using misinformation as an umbrella term for any false or misleading content regardless of the user's intention \cite[e.g.,][]{johansson2022can,pantazi2021social}.

With the spread of user-generated content and social media platforms, there has been a marked increase in scholarship about misinformation online \cite{altay2023misinformation}. Over the same period there has been an increase in fact-checking organizations: teams of journalists aiming to fact-check or debunk misinformation \cite[e.g.,][]{graves2016rise, siwakoti2021covid}. Meta's third-party fact-checking program pays organizations to write fact-checks about content on its Facebook and Instagram platforms \cite{metafactcheck} or to host `tiplines' on WhatsApp where users can search fact-checks \cite{johansson2022can, kazemi2022research}.
Google also collects copies of fact-checks in ClaimReview markup for use in its news and search tools \cite{factcheckgoogleclaim}. Much of the data Google collects is freely available and is one of our data sources for this study. 

Fact-checking organizations typically follow similar principles in their work, aiming to verify claims from across the political spectrum, relying on authoritative data and independent experts, and striving for transparency in their methods and funding sources \cite{graves2016rise}. However, specific practices can vary. In selecting claims to check, some organizations take a systematic approach to maintain balance, such as fact-checking every claim made in political debates, while others focus on viral misinformation or statements by prominent political figures \cite{graves2016rise}. The decision-making process often involves editorial judgement about which claims are most consequential or in need of verification. Otherwise, fact-checkers may struggle to respond to shifting political circumstances and focusing on the most impactful claims \cite{graves2016rise}.

Fact-checking organizations operate under various models, with some affiliated with larger media organizations, universities, or functioning independently \cite{slijepvcevic2021media}. Many fact-checking organizations are small non-profits, and monetization remains a significant challenge. While some benefit from the resources of established media companies, independent fact-checkers often struggle with sustainable funding \cite{graves2016rise}. Financing sources include charitable foundations, individual donations, and in some cases, partnerships with tech companies like Meta and Google \cite{slijepvcevic2021media,lelo2022rise}. A few outlets have managed to generate revenue through broadcast partnerships or by providing professional services and training \cite{graves2016rise}.

Many fact-checking organizations are signatories to the IFCN Code of Principles, which outlines standards for non-partisanship, fairness, transparency, and accountability \cite{ifcn2024signatories}. Indeed, being a signatory to the IFCN code of principles is required to participate in Meta's third-party fact-checking program.

Fact-checking resources are unevenly distributed across the globe. North America, Europe, and Australia have more fact-checking organizations than other regions of the world, although the difference is decreasing \cite{stencelMisinformation2023}. English remains the most-resourced language: 33.1\% of fact-checks in our dataset are English. This stands in stark contrast to the global distribution of people and Internet users in the world \cite{graham2012featured}.

\subsection{(Mis)information across languages}\label{sec2-MisinfoCross}
Most research on misinformation diffusion has not explicitly considered cross-language spread but more general research has found geography, language, and culture, slow the spread of information \cite[e.g.,][]{barnettPhysical1995,haleImpact2012,haleNet2012,haleMultilinguals2014,takhteyevGeography2012}. According to the culture proximity theory, people tend to prefer content that is most proximate to their location, language and culture \cite{straubhaarMedia1991}. Language stands as a salient explanation for the cultural proximity in information consumption, and scholars have identified several major online content consumption clusters \cite{ksiazekCultural2008}. Rather than people consuming content from the entirety of the \emph{World Wide} Web, audiences tend to consume information in their preferred languages \cite[e.g.,][]{ksiazekCultural2008,curtinMedia2003,straubhaarWorld2007}. 

However, the development of social media technology and global events (e.g., pandemics, conflicts, and climate change) may reduce language barriers, facilitating a higher percentage of cross-language diffusion than before \cite{zuckermanRewire2013}. In turn, there may now be a higher percentage of common misinformation across languages as shared social media platforms and translation technology as well as bilingual users make it possible to consume more diverse content \cite{zuckermanRewire2013, haleGlobal2014}. As noted, COVID-19 and the Russian invasion of Ukraine have formed the basis of new fact-checker collaborations (\#CoronaVirusFacts and \#UkraineFacts).

A variety of studies have investigated multilingual misinformation. The majority of these studies focused on a single topic, considered only a small number of languages, or investigated misinformation on a single platform.\citet{madraki2021characterizing} study COVID-19 misinformation ``across languages, countries and platforms'' grouping misinformation into topics, and analysing their prevalence with respect to different domains and in three languages (Chinese, English, and Farsi). The authors find that politics is at the root of most of the collected misinformation across all three languages. \citet{shahi2020fakecovid} present a first multilingual and cross-domain dataset comprising of 5,182 fact-checked news articles. They manually annotated fact-checked articles into 11 different categories and performed explanatory analyses of their contents. They acknowledge repeated misinformation while building a classifier for misinformation about Covid-19 and performed an exploratory analysis of their contents. \citet{li2020toward} construct a multimodal ``fake-news data repository'' focusing on Covid-19 in six different languages. \citet{nielsen2022mumin} build a multilingual multimodal fact-checked misinformation dataset comprising of 13 thousand fact-checked tweets in over forty languages. The paper proposed two classification tasks where social media elements (tweets, users, images, claims) are represented as nodes to ascertain claim veracity, but does not analyze how misinformation spreads across different languages. 

We are unaware of any large-scale studies examining the spread of repeated misinformation across topics and languages globally. Our study is unique in the breadth of the dataset considered, and differs from previous studies by examining how repeated misinformation is fact-checked, spreads, and evolves. 

\section{Data}\label{sec3}
The dataset used in this study is a combination of data from Google Fact-Check explorer \cite{DataCommons2023} and data directly crawled from the websites of verified signatories of the International Fact-Checking Network (IFCN) code of principles. We find that Google's data is incomplete: it only includes fact-checks with structured ClaimReview metadata, but not all fact-checkers include this data. We wrote custom crawlers in June 2020 for fact-checking websites that we identified were missing from the Google data. We gathered all available historic articles from those websites at that time. We then ran hourly crawls to gather any newly published content through the end of the study. We structure crawled pages in ClaimReview format to match the format of the Google data.\footnote{Our code, including a list of the websites crawled, is available open-source at \url{https://github.com/meedan/fetch}.}

ClaimReview markup consists of multiple data fields. First, each fact-check has an associated author and date. Additionally, it contains the \emph{Claim}---that is the statement of misinformation---that is being fact-checked, a \emph{Review}, and a \emph{Rating}. The Google Fact-Check Explorer data contains 66,566 unique fact-checks, and our crawled data contains 264,487 unique observations. We combined these two datasets to generate the final dataset, and deduplication and data cleaning steps are outlined in Section \ref{DataPrep}. 
We limit our analysis to the period of time from January 2020 to January 2023 as the amount of fact-checks present in our dataset is not stable outside of this time frame leading to much lower estimates of the number of clustered claims. 

\subsection{Data Preparation} \label{DataPrep}
Not all fact-checkers adhere completely to the ClaimReview format. When the \texttt{claimReviewed} field is missing or empty, we consider the \texttt{headline} and \texttt{description} fields. We employ a heuristic to determine which part of the fact-check contains the claim: If the fact-checking entry contain any string in the \texttt{claimReviewed} field, we return this unaltered. Otherwise, we check whether either the length of the \texttt{headline} or the \texttt{description} fields is within two standard deviations of the average length of a \texttt{claimReviewed} entry. If this is the case, we return either one (with preference for the former). If both are longer we remove the fact-check.

Similarly, not all fact-checks contain the name of the organization posting the claim. To remedy this, we extract the domain name of the final redirect of each URL associated with a fact-check and use this as our primary entity identifier. Another pre-processing step taken to assure that only the claim is contained in the final dataset, we manually inspect tri- to six-grams within each fact-checking domain that appear disproportionally often. We used the LaBSE tokenizer to split each fact-check, and aggregate token counts by the domain of the fact-checker. Examples of tokens, unrelated to the actual claim are ``WHATSAPP - CHECK," ``Verificamos," ``Fact-Check:". After manually reviewing the most repeated substrings, we found 84 that are not part of the claims being fact-checked and remove these. In addition, we remove exact duplicates of fact-checks (retaining the earlier in time), and fact-checks that did not contain a valid claim, according to the aforementioned heuristics employed, after the pre-processing. 

Lastly, the initial data contained a significant number of fact-checks that appear to only be editorial mistakes. These fact-checks only differ based on punctuation, or slight editorial fixes, and were posted extremely close together in time. To remove these fact-checks we check for duplicates after removing any punctuation and non-alpha-numeric characters and ignoring case. Additionally, we remove any fact-checks that have a cosine similarity exceeding 0.95 measured with LaBSE and are posted by the same domain or author. 

\subsection{Final Dataset}
Our final dataset consists of 264,487 unique fact-checks. Each fact-check has an associated claim, verdict (or rating), date, and author. The language of each claim was determined using the Google Translate API: we found a total of 95 unique languages, showcasing the diversity of fact-checking organizations dedicated to improve the information environment worldwide.

\begin{figure}[tb]
\centering
\includegraphics[width=\columnwidth]{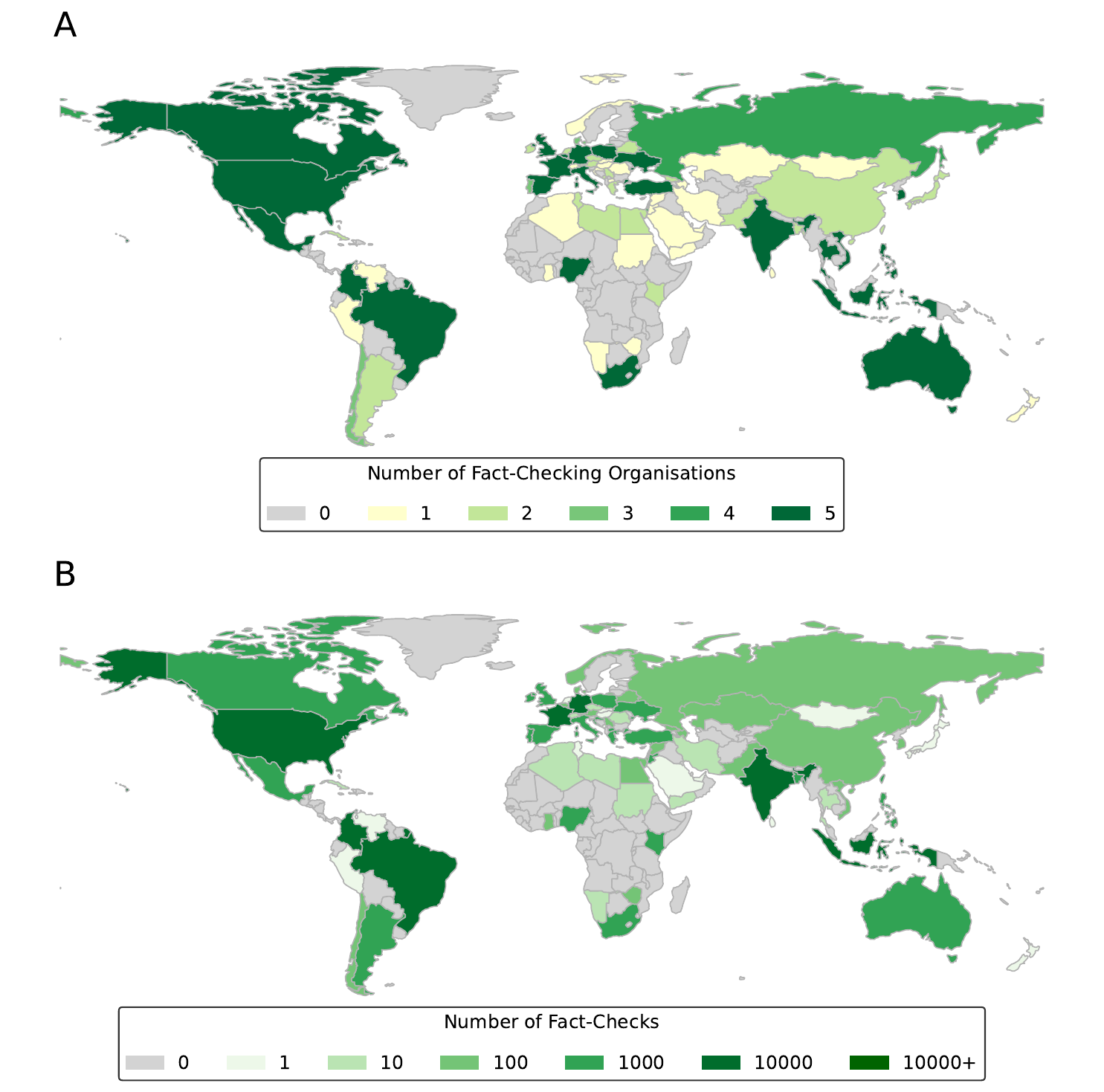}
\caption{Geographical Distribution of Fact-checking Organizations and Fact-checks. \textbf{A} shows the number of Fact-checking organisations in our dataset by country. \textbf{B} shows the number of fact-checks published by fact-checking organisations from each country.}
\label{map}
\end{figure}

Figure \ref{map} shows the number of fact-checking organizations contained in our dataset and their respective countries of origin. We extracted the country of origin for all fact-checking organizations with at least ten claims in our dataset from the IFCN website, and, where not available, through manual searches on each website. The Figure does not include organizations which do not have a clear country of origin (e.g., Africa Check), or focus on an entire region/continent. Panel A highlights all countries based on the number of fact-checking organizations, while panel B shows the number of fact-checks by IFCN signatories contained in our dataset. The map demonstrates that fact-checking is a global activity.
To further illustrate the linguistic composition of our dataset, Table \ref{tab:language_distribution} presents the distribution of fact-checks across different languages.

\begin{table}[h]
    \centering
    \begin{tabular}{ccc}
    \toprule
    Language & Number of Fact-checks & Percentage \\
    \midrule
    English & 87,536 & 33.10 \\
    Portuguese & 37,673 & 14.24 \\
    Spanish & 23,209 & 8.78 \\
    Hindi & 17,491 & 6.61 \\
    Arabic & 12,992 & 4.91 \\
    Indonesian & 10,414 & 3.94 \\
    Bengali & 8,659 & 3.27 \\
    German & 7,540 & 2.85 \\
    French & 5,970 & 2.26 \\
    Turkish & 5,948 & 2.25 \\
    \midrule
    Other & 47,055 & 17.79 \\
    \bottomrule
    \end{tabular}
    \caption{Distribution of Fact-checks by Language.}
    \label{tab:language_distribution}
\end{table}

As shown in Table \ref{tab:language_distribution}, English accounts for the largest proportion of fact-checks (33.10\%), followed by Portuguese (14.24\%) and Spanish (8.78\%). However, it's noteworthy that non-English fact-checks collectively represent about two-thirds of the dataset, underscoring the linguistic diversity of fact-checking efforts globally. The breadth of the linguistic and geographic diversity of fact-checking organizations highlights the need for research into multilingual misinformation, extending beyond a focus on European languages and Western cultural and political contexts. While English remains the most prevalent language, the significant presence of fact-checks in languages such as Portuguese, Spanish, Hindi, and Arabic demonstrates the global nature of both misinformation spread and fact-checking efforts.

\begin{figure}[tb]
\centering
\includegraphics[width=\columnwidth]{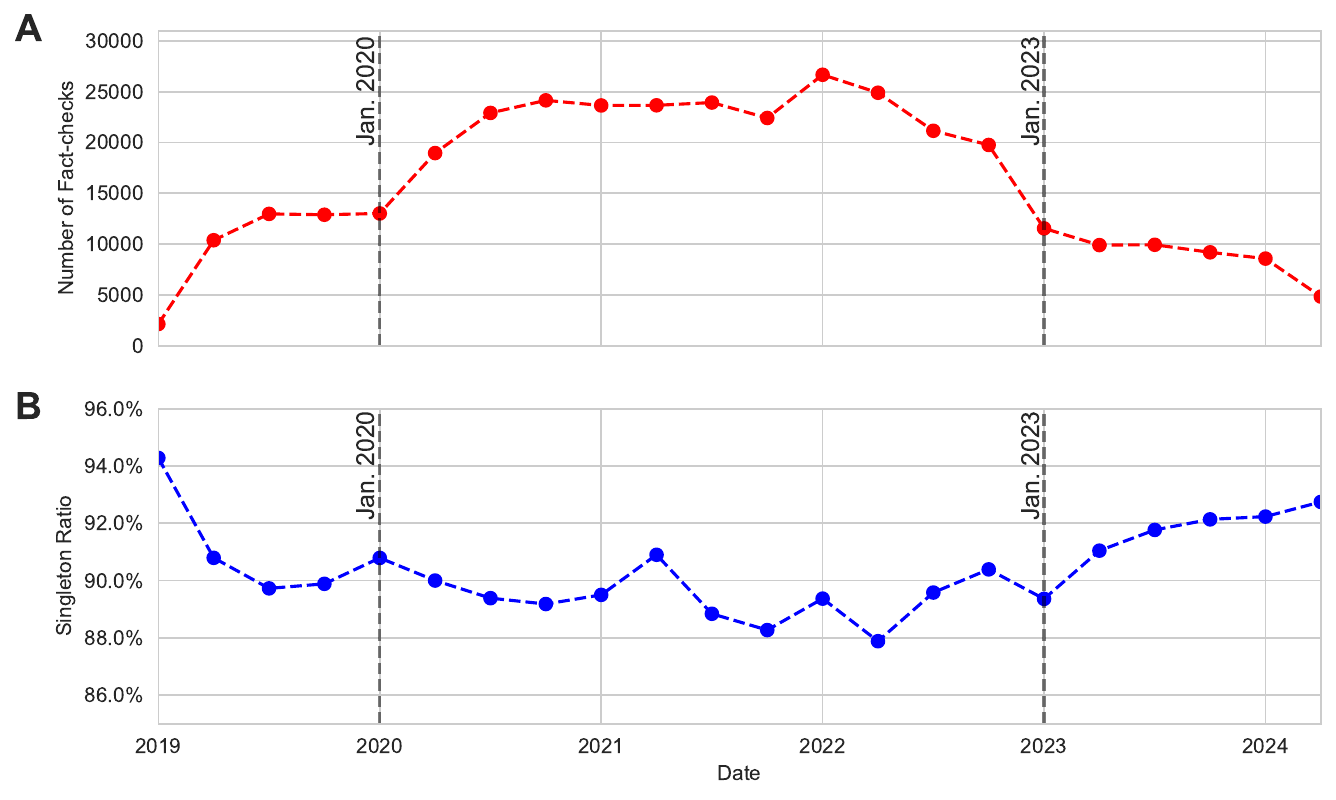}
\caption{Number of Fact-checks in Dataset \& Percentage of unique Claims. Timeframe between vertical lines indicates data included in the analysis. \textbf{A} Number of fact-checks in the dataset. The timeframe between the vertical, dashed lines indicate the analysis period. Fact-checks outside of this period are excluded to assure a consistent number of data-points.  \textbf{B} Percentage of fact-checks that are not clustered with any other fact-check, averaged over three month periods. See Section \ref{sec:ClustEval} for a comprehensive discussion of the clustering methodology.}
\label{weeklycounts}
\end{figure}

The dataset spans from 2018 to 2024, and covers a broad array of topics. Panel A of Figure \ref{weeklycounts} displays the total number of fact-checks per three-month period in the dataset. We observe that the number of fact-checks is relatively consistent between January 2020 and January 2023 at twenty thousand fact-checks per three month interval. Before this period, we see around four thousand fact-checks per month and after it around one thousand. Due to this variation in fact-check volume, we restrict our analysis to the period between January 2020 and January 2023 where the data coverage is most complete and consistent. In panel B, we show the percentage of the claims that are ``unique." A thorough discussion of the definition of a unique claim will follow in Section \ref{sec:Methods}. In the periods before and after the observations period (January 2020--January 2023), the percentage of unique fact-checks is significantly distorted due to the lower number of available fact-checks. We therefore restrict ourselves to this time period. We additionally, analyze the prevalence of fact-checks related to two major events (Covid-19 and the invasion of Ukraine) as these topics may be large enough to significantly alter the results of our study. We find that while both topics make up a significant proportion of the fact-checks in the analysis, their prevalence is not sufficient to change any of the conclusions of the article. The exclusion of both topics reduces the percentage of non-clustered claims from 10.26\% to 9.46\% and the percentage of clustered claims that are multilingual from 32.26\% to 30.63\%. The results of the analysis are detailed in Section \ref{Appendix:A} of the appendix.

\section{Methods} \label{sec:Methods}
To compare misinformation spread across languages, we embedded all fact-checks with Language-agnostic BERT Sentence Embedder (LaBSE) \cite{feng-etal-2022-language}. LaBSE combines Masked Language Modeling (MLM) where a representation is learned by randomly masking tokens in one language, and letting the model predict the token, and Translation Language Modeling (TLM) where bi- or multilingual sentences are concatenated words are masked in both sentences. The model then predicts the masked words, encouraging it to learn cross-lingual representations. In contrast to other multilingual sentence embedding models, LaBSE supports at least 109 languages, enabling us to include a larger number of fact-checks, even in languages typically considered low-resource. LaBSE is therefore ideal to retrieve similar sentences across languages. Nevertheless, as a robustness check we embedded all covered fact-checks with distiluse-base-multilingual-cased-v2, paraphrase-multilingual-MiniLM-L12-v2, and paraphrase-multilingual-mpnet-base-v2 \cite{reimers-2019-sentence-bert}. 
The results were qualitatively similar regardless of the embedding model used. We proceed with LaBSE as its 109 languages cover 99.23\% of our data.

To cluster the fact-checking claims we utilize the LaBSE embeddings and retrieve other fact-checks with a cosine similarity exceeding a given threshold. To retrieve the approximate nearest neighbors we employed Locality Sensitive Hashing (LSH), specifically Spotify's ANNOY library \cite{spotannoy}, to reduce the number of computations. We used 100 hyperplanes to retrieve the nearest neighbors. To gather the approximate nearest neighbors, we started by retrieving 10 nodes (i.e., embedded fact-check claims), and doubled the number of retrieved claims until the last element of the retrieved neighbors fell below the cosine similarity threshold. We then performed a binary search within the last batch of returned nodes to determine the last element to be included. The resulting data-structure can be modeled as an extremely sparse graph. We then extracted all connected components to yield the final set of 
clusters. Our methodology to embed fact-checks, retrieving the approximate nearest neighbors with a cosine similarity surpassing a given threshold, and subsequently extracting connected components, has proven particularly potent when grappling with the dimensionality of LaBSE embeddings and the sparse nature of the resulting graph. Contrasting this with other clustering methods, density-based approaches such as HDBScan \cite{mcinnes2017} or DBSCAN \cite{ester1996} offer advantages in terms of identifying clusters of various shapes and densities and have robust noise-handling capabilities. However, they are computationally intensive and struggle to define density in high-dimensional, sparse spaces, making them less ideal for sentence embeddings. Centroid-based clustering methods, like k-means \cite{macqueen1967}, though widely adopted, can be adversely affected by noise and outliers, and assume clusters to be convex-shaped, which may not hold true in our context. By comparison, our proposed strategy balances computational efficiency and robustness to noise. The use of a preset cosine similarity threshold facilitates control over cluster granularity, and the extraction of connected components naturally separates noise and outliers into distinct clusters. This combination of strategies results in an efficient, interpretable, and scalable solution to the problem of clustering a large number of multilingual sentence embeddings.

\begin{figure}[tb]
\centering
\includegraphics[width=1.1\columnwidth,height=20cm,keepaspectratio]{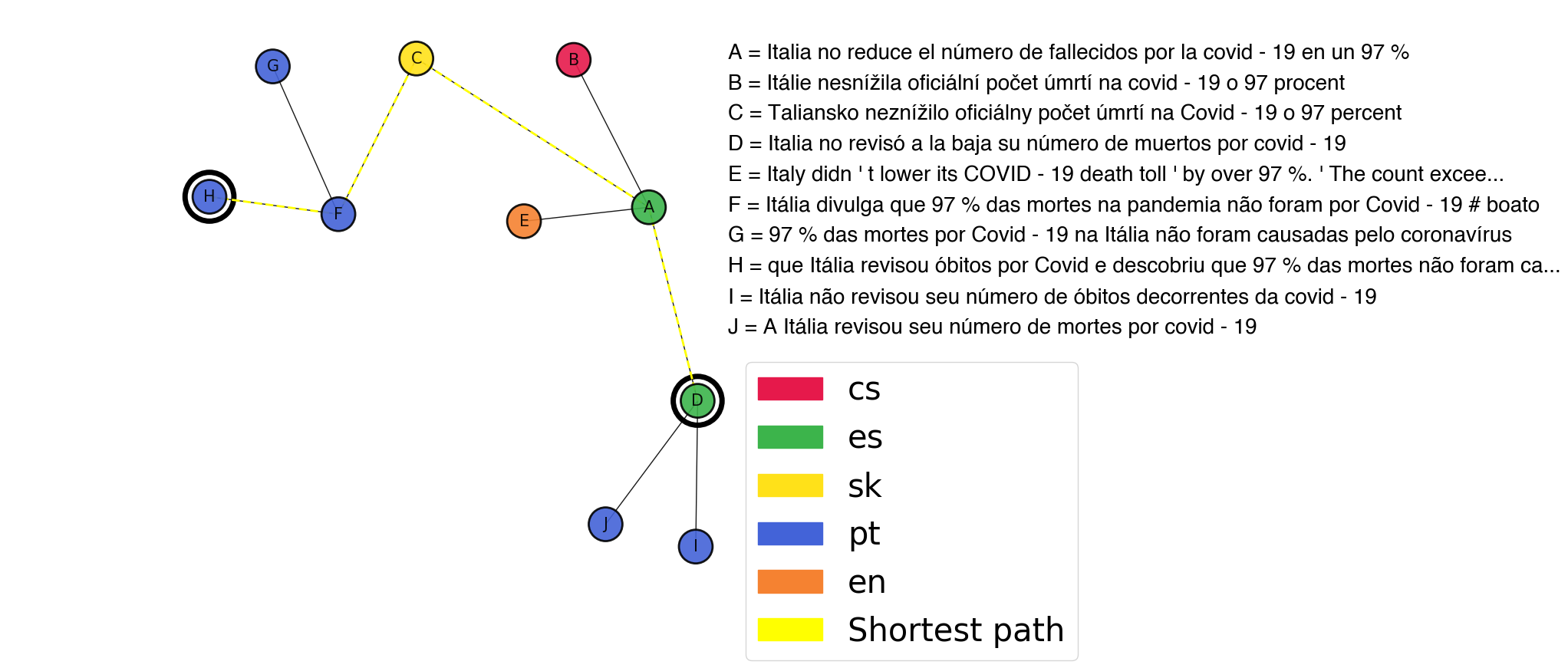}
\caption{Exemplary Multilingual Cluster with Shortest Path. The two most dissimilar nodes are circled in black, the shortest path between them is highlighted in yellow.}
\label{exampleCluster}
\end{figure}

Figure \ref{exampleCluster} displays an example cluster of ten fact-checks authored in six different languages. The embeddings of the claims are projected on two dimensions with Uniform Manifold Approximation and Projection (UMAP) \cite{mcinnes2020umap}. The two most dissimilar nodes are circled in black. The shortest path between the two nodes is highlighted in yellow. 

\subsection{Clustering Evaluation} \label{sec:ClustEval}
To evaluate the robustness, accuracy, and consistency of the clustering, we went through four rounds of human validation to test the performance of different clustering thresholds. We randomly sampled 100 clusters each time and asked the expert coders to qualitatively code the most dissimilar pair as indicated by the cosine similarity and evaluated whether they were talking about the same misinformation claim. The coding scheme follows the work of \citet{kazemiClaim2021}. To ensure consistency across the evaluation process, we employed the same two expert coders for all four rounds of validation. Both coders were doctoral students, familiar with fact-checking practices, and had experience in content analysis. They worked independently to evaluate the clusters, and any disagreements were resolved through discussion to reach a consensus. The coding sheet provided to the coders included machine translations of each claim to facilitate understanding. The first three rounds of human evaluation identified the need for better pre-processing to remove strings from claims such as publishers' names, special characters, and formatted fact-checking claims.

\begin{figure}[tb]
    \centering
    \includegraphics[width=\columnwidth]{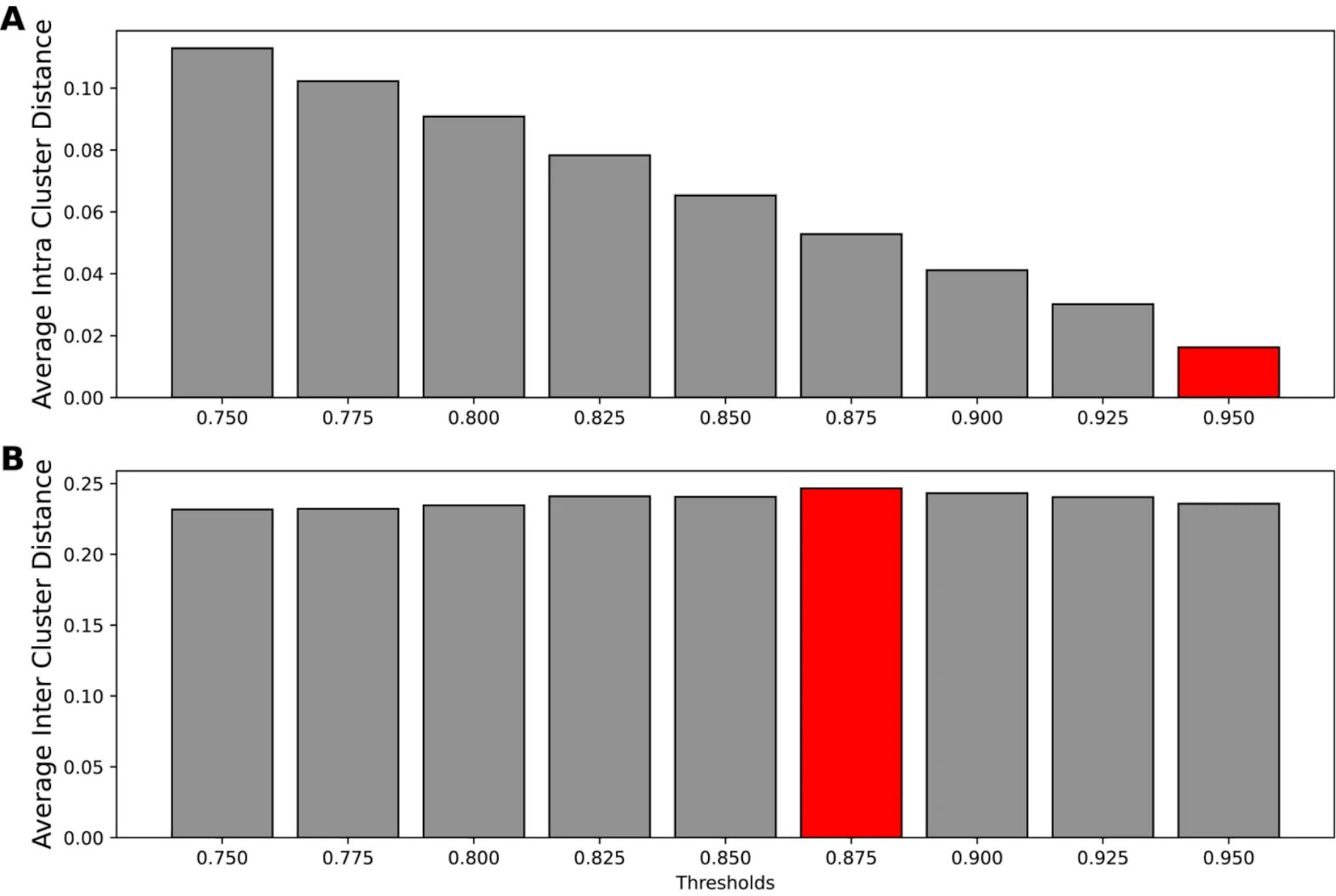} 
    \caption{\textbf{A} Average \emph{intra}-cluster cosine distance by threshold. The inter-cluster distance decreases with an increasing threshold. \textbf{B} Average \emph{inter}-cluster cosine distances by threshold. Highlighted bars show the minimized intra-cluster variance and the maximized inter-cluster distance.}
\label{fig:threshold-analysis}
\end{figure}

In addition to the qualitative analysis, we examine three measures of the goodness of fit of the clustering. First, we analyze the average of cosine distance within clusters (intra-cluster distance). Better clusters should lead to a reduced variation for higher thresholds. Additionally, we looked at the between-cluster (or inter-cluster) cosine distances. Here we selected the centroid of a cluster and sampled up to 10,000 additional centroids for which we then calculated the average cosine distance. The mean  \emph{intra}-cluster variance and \emph{inter}-cluster distance are shown in Figure \ref{fig:threshold-analysis} for different thresholds.

The mean intra-cluster variance consistently falls with higher thresholds (Figure \ref{fig:threshold-analysis} top). This is a positive result because it signifies that our clusters are becoming more cohesive---the elements within each cluster are more alike. This increased cohesiveness is crucial for our clustering methodology as we aim to gather similar fact-checks together to facilitate their analysis. The highlighted bar shows the cosine similarity threshold for which the intra-cluster variance is minimized. 

Conversely, the inter-cluster distance is the average distance of randomly chosen cluster pairs for each threshold. An increasing inter-cluster distance signifies that---as we increase the cosine similarity threshold---the clusters are becoming more distinct and there is better separation \emph{between clusters}. The highlighted bar panel B of Figure \ref{fig:threshold-analysis} shows inter-cluster distance is maximized at 0.875.

While intra-cluster variance is minimized with a cosine similarity threshold of 0.95---the maximum similarity we tested---the inter-cluster distance is maximized with a threshold of 0.875. Increasing the cosine similarity beyond this point leads to clusters that are closely linked, to be split into two components, thereby reducing the inter-cluster distance. We also find larger thresholds yield smaller sized clusters. At a cosine similarity threshold of 0.875, the average non-singleton cluster size is 2.59. This contrasts with average cluster sizes of 6.53 and 2.13 at thresholds of 0.75 and 0.95, respectively.

Lastly, another heuristic measurement available to use is the rating associated with each fact-check (e.g., `True,' `False,' `Misleading,' etc.). As different media organizations employ different rating schemes, we first removed potentially confounding spellings, white-spaces, punctuation, and case from each verdict (or rating). We then selected any assessment that appeared at least 50 times in the dataset and manually mapped these to ``false," ``mostly-false," ``mostly-true," and ``true." For this evaluation of clusters, we ignored the 30\% of fact-checks with ratings not in this mapping.

\begin{table}
\centering
\begin{tabular}{@{\extracolsep{5pt}} lcccccc}
\toprule
Threshold & 0.8 & 0.825 & 0.85 & 0.875 & 0.9 & 0.925 \\ 
\midrule
4-Cat. Majority & 86.76 & 89.92 & 91.78 & 93.07 & 94.38 & 95.54 \\
& (75.73) & (76.32) & (76.61) & (77.64) & (77.54) & (78.29) \\
\midrule[\heavyrulewidth]
2-Cat. Majority & 97.27 & 97.84 & 98.25 & 98.52 & 98.83 & 99.06  \\
 & (92.96) & (93.12) & (93.05) & (93.10) & (93.06) & (93.14)  \\
\bottomrule
\end{tabular}
\caption{Evaluation of clustering goodness of fit by comparing the consistency of the fact-check verdicts/ratings of claims in each cluster. Ratings are mapped into the two categories true and false (\emph{2-Cat. Majority}) or the four categories true, mostly-true, mostly-false, and false (\emph{4-Cat. Majority}). Below each statistic, a random baseline is given. The random baseline is calculated by sampling from the verdict distribution and randomly assigning labels to each node in a connected-component. In all cases, the actual mapping is significantly more accurate than the random baseline.}
\label{ClusteringGoodness}
\end{table}

Having fact-checks in the same cluster generally sharing the same rating is an important signal of the correctness of our clusters. Table \ref{ClusteringGoodness} shows two measures of the clustering accuracy. We show the percentage of claims for each cluster in the modal verdict category. In the first row, the values indicate the percentage of claims in the modal category for four different categories (true, mostly-true, mostly-false, and false). The values in parenthesis indicate a random baseline. The second row of the table shows the percentage of claims per cluster in the modal category mapped only to true and false. For any cosine-similarity higher than 0.825 we find that over 90\% of fact-checks in the same cluster have the same associated rating. This holds both when we employ two labels (true, false) and when we employ four labels (true, mostly-true, mostly-false, false). This indicates that our clustering methodology effectively groups fact-checks that are not only semantically similar but also share similar ratings of truthfulness. In all experimental conditions, our clustering outperforms a random baseline, constructed by randomly sampling verdicts from the overall verdict distribution. 

We finally settled on a cosine-similarity threshold for edges of 0.875 after analyzing all available measures in conjunction with our qualitative evaluation. This measure is in line with previous research \cite[e.g.,][]{kazemiClaim2021}. Where possible we perform measures with a range of thresholds, and we find the same general patterns.

With a threshold of 0.875, our qualitative analysis showed 93 of the 100 pairs of most dissimilar nodes we sampled from clusters belonged to the same cluster. Overall, we have high confidence in the precision of our clustering methods from the qualitative analysis, consistency of ratings, and high intra-cluster similarity. Similarly, the low inter-cluster similarity suggests recall is generally good; nonetheless, there could be instances where we fail to identify two claims as similar.

\section{Results}
\subsection{Research Question 1: To what extent are misinformation claims fact-checked by multiple fact-checking organizations?}
To answer \emph{RQ1} we extract the connected components of our sparse graph. Any component (or cluster) with two or more nodes (fact-checks) contains a claim that has been fact-checked at least twice. The remaining nodes are singletons, and have not been matched with any other fact-check. The proportion of singleton nodes ranges from 73.8\% for a cosine similarity threshold of 0.8 to 93.5\% for a threshold of 0.9. For the 0.875 threshold we find that 89.7\% of nodes are singletons---or conversely---that 10.26\% of fact-checks investigate a claim that has previously been fact-checked. This represents a significant proportion of repeated work across fact-checking organizations investigating the same claims. 

\subsection{Research Question 2: What percentage of non-unique fact-checks spreads across languages}
To address our second research question, we quantitatively investigate the spread of non-unique fact-checks across languages. From the claims that are fact-checked more than once, we find that 32.26\% are fact-checked in multiple languages, suggesting the original misinformation claim was present in multiple languages. Figure \ref{weeklycounts} shows the singleton ratio of the entire dataset, including the discarded time-periods, over time. This finding agrees with previous studies, showing that misinformation does not exist in language-specific isolation but rather has the potential to traverse language barriers \cite{zeng2021cross, kazemiClaim2021}.

\begin{figure}[tb]
\centering
\includegraphics[width=\columnwidth]{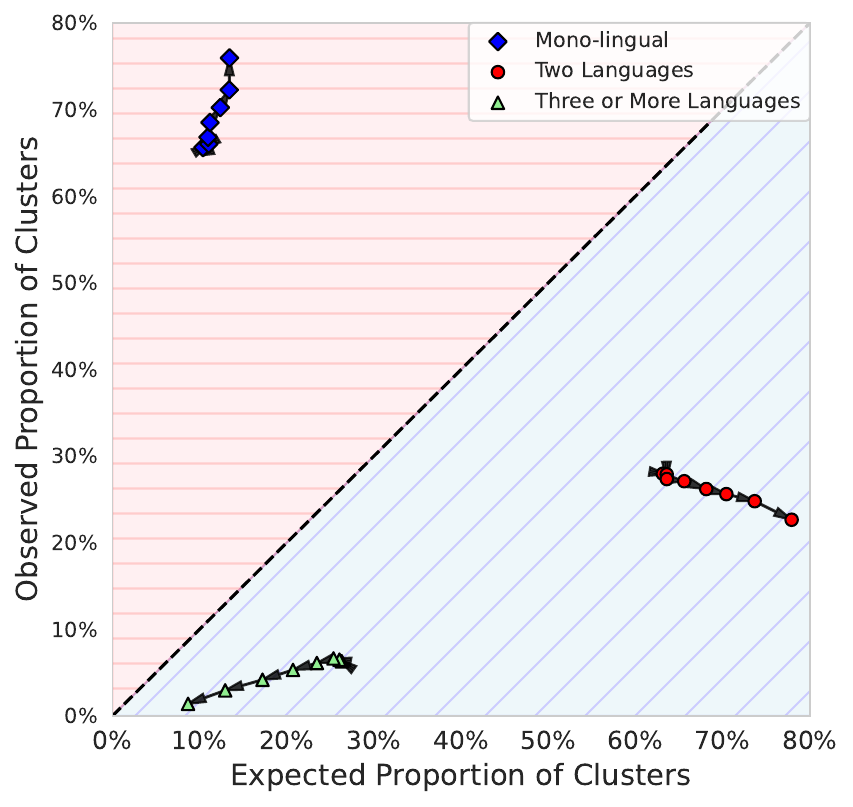}
\caption{Experimental Evaluation of Language Assortativity: Different thresholds are shown for each group, and arrows show the direction with an increasing threshold. The color and the shape of the nodes indicates the type of cluster. Scatters on the 45-degree line would indicate equal observed and expected frequencies. Monolingual clusters are observed more than expected when compared to the random baseline, while bi- and tri-lingual clusters are observed less than expected.}
\label{ActVsRand}
\end{figure}

Despite this, our research also indicates an inclination for misinformation to disseminate predominantly within the same language. To substantiate this, we compare our observed data against a null model, which assumes no language-based preferences for misinformation spread. The null model's expectations are calculated by randomly sampling languages from the overall language distribution but keeping all edges of the graph unchanged. To clarify the concept of randomness in our null model, we assume that the spread of misinformation is not influenced by language barriers, behaving as if language preference does not exist. As the languages associated with misinformation claims are chosen by random sampling from the overall language distribution, we nullify any inherent language-based patterns or preferences. Comparing to this null model as a baseline allows us to understand how language could specifically impact the dissemination of misinformation. We depict this analysis in Figure \ref{ActVsRand}, where we plot the expected frequencies from the null model against the empirically observed frequencies of mono-, bi-, and tri-or-more- lingual clusters. Each point corresponds to a distinct cosine similarity threshold. 
If there were no language-specific effects in the clustering, we'd expect the points to lie on the 45-degree line $y=x$ as we'd observe the value just as often as is expected. The empirical data deviates significantly ($\alpha = 0.01$) confirming the influence of language on our clustering of misinformation claims.

To further understand the geographical dynamics of fact-checking efforts, we analyzed the distribution of multiple fact-checks within and across countries. We extracted the country of origin for all fact-checking organizations with at least ten claims in our dataset, and where not available, through manual searches on each website. Where we were unable to find countries of origin, we relied on domain suffix identifiers. Our analysis reveals that while 32.26\% of repeated fact-checks cross language barriers, the proportion of fact-checks that cross country barriers is lower at only 20.68\%. To investigate the extent of collaboration within countries, we examined claims that were repeatedly fact-checked within the same country but by different organizations. We found that out of 23,134 fact-checks addressing claims fact-checked more than once in the same country, 57.41\% were fact-checked by only a single organization.

To investigate what type of claims are most likely to be fact-checked multiple times, spread across languages, or re-occur, we first machine translate all claims to English using Google Translate. We then extract and lemmatize all noun tokens from the claim fields. Nouns are identified using NLTK's part-of-speech tagging \cite{hardeniya2016natural}. Subsequently, we calculated the relative frequency of each token under three conditions. First, whether the claim is a singleton (cluster size = 1) or not (cluster size $>$ 1). Secondly, for non-singleton claims, whether the claim cluster is monolingual or multilingual.\footnote{We look at the original dataset before machine translation to determine whether a cluster is mono- or multilingual.} Lastly, we calculate the range of publishing dates for the fact-checks by cluster. We designate clusters with fact-checks published more than a month apart as ``long-lasting''. The median value for the range of the publishing dates of fact-checks by cluster is 4 days. Our chosen value of 30 days approximately coincides with the third quartile of the distribution. Before calculating the relative frequencies, we filtered the lists to tokens present in both conditions. When calculating the relative frequencies, we only included tokens that appeared at least 50 times.

\begin{table}[tb]
\centering
\begin{tabular}{@{\extracolsep{5pt}} cccccc} 
\toprule
\multicolumn{2}{c}{Multilingual} & \multicolumn{2}{c}{Multiple Fact-Checks} & \multicolumn{2}{c}{Long Lasting}\\
\cmidrule(r){1-2}\cmidrule(l){3-4}\cmidrule(l){5-6}
Multilingual & Rel. Frequency & Clustered & Rel. Frequency & Long Lasting & Rel. Frequency\\
\cmidrule(r){1-1}\cmidrule(lr){2-2}\cmidrule(lr){3-3}\cmidrule(l){4-4}\cmidrule(l){5-5}\cmidrule(l){6-6}
name & 3.33 & greta & 6.84 & ivermectin & 3.22 \\
pfizer & 2.99 & thunberg & 6.68 & candidate & 3.10 \\
rahul & 2.64 & ipec & 3.59 & covid19 & 2.63 \\
virus & 2.42 & shahrukh & 3.40 & risk & 2.49 \\
delhi & 2.36 & boom & 3.22 & cure & 2.48 \\
george & 2.27 & ivermectin & 2.99 & test & 2.48 \\
ivermectin & 2.23 & ceo & 2.84 & covid & 2.46 \\
covid19 & 2.18 & dilma & 2.76 & disease & 2.34 \\
gandhi & 2.16 & sonia & 2.65 & blood & 2.33 \\
khan & 1.93 & soros & 2.49 & list & 2.22 \\
... & ... & ... & ... & ... & ...\\
man & 0.70 & party & 0.41 & woman & 0.63 \\
time & 0.70 & city & 0.42 & montage & 0.61 \\
number & 0.68 & facebook & 0.42 & trump & 0.59 \\
show & 0.67 & university & 0.43 & tweet & 0.59 \\
case & 0.65 & lockdown & 0.44 & bjp & 0.58 \\
country & 0.64 & bank & 0.45 & leader & 0.55 \\
rumor & 0.58 & school & 0.48 & post & 0.54 \\
montage & 0.54 & plan & 0.48 & hospital & 0.51 \\ 
government & 0.44 & truth & 0.48 & vote & 0.44 \\
vote & 0.38 & internet & 0.49 & police & 0.43 \\
\bottomrule
\end{tabular}
\caption{Tokens Most and Least Associated with Multilingual Clusters, Clusters with Multiple Fact-Checks or Long-Lasting Clusters.  Relative frequencies are calculated by dividing the relative frequency in both conditions.}
\label{multiclaimtab}
\end{table}

Table \ref{multiclaimtab} presents tokens that are most and least likely to appear in multilingual clusters (left column), in clusters containing multiple fact-checks (center column), and ``long lasting'' clusters (right column).

For instance, a token indicative of multilingual clusters is ``Pfizer'', emphasizing that Covid-19 is a globally discussed topic. It has a relative frequency of 2.99 meaning that it occurs close to three times more often in clusters which contain more than one language than in mono-lingual clusters. Other top multilingual tokens like ``virus'' also relate to the pandemic. Conversely, tokens like ``case,'' ``country,'', ``government'', and ``vote'' are primarily confined to single-language discourse. The latter relate to domestic politics and elections, which are unlikely to be fact-checked in more than one language. Tokens such as ``Ivermectin,'' ``Soros,'' and ``Greta Thunberg'' are often found in claims checked multiple times. These terms relate to global topics, but are also conspiratorial in nature. In contrast, ``party'', ``city'', and ``university'' are less associated with viral conspiracy theories. The tokens that are most predictive of ``long-lasting'' claims are wholly related to Covid-19 and conspiracy theories relating to the disease.

This exploratory analysis shows that there are significant topical differences in the type of claims that are fact-checked more than once, spread across languages, and lost longer than one month. Tokens that appear to be related to Covid-19 conspiracy theories dominate our analysis of tokens most associated with multilingual clusters. Lastly, this analysis indicates that there are repeated tropes and narrative themes that are most likely to be fact-checked more than once.

\subsection{Research Question 3: What languages share the most misinformation?}

To answer our third research question asking which languages have the most misinformation in common, we analyze the language pairs and triples that most frequently co-occur in our clustered fact-checks. As shown in Table \ref{tab:language_switches}, the most common language pair for common misinformation is Spanish and Portuguese, with 362 instances. This is followed by English and Hindi (294 instances), and Hindi and Punjabi (also 294 instances). Other frequent pairs include Bengali and English (237 instances), and Hindi and Urdu (181 instances). When considering triples of languages, we find that Hindi, Punjabi, and Urdu form the most common group with 63 instances, followed by English, Hindi, and Punjabi (27 instances), and Bengali, English, and Hindi (23 instances).

\begin{table}[h]
\centering
\begin{tabular}{crcr}
\hline
\textbf{Language Pairs} & & \textbf{Language Triples} & \\
\hline
Spanish $\leftrightarrow$  Portuguese & 362 & Hindi $\leftrightarrow$ Punjabi $\leftrightarrow$ Urdu & 63 \\
English $\leftrightarrow$ Hindi & 294 & English $\leftrightarrow$ Hindi $\leftrightarrow$ Punjabi & 27 \\
Hindi $\leftrightarrow$ Punjabi & 294 & Bengali $\leftrightarrow$ English $\leftrightarrow$ Hindi & 23 \\
Bengali $\leftrightarrow$ English & 237 & English $\leftrightarrow$ Indonesian $\leftrightarrow$ Malay & 14 \\
Hindi $\leftrightarrow$ Urdu & 181 & English $\leftrightarrow$ Hindi $\leftrightarrow$ Urdu & 14 \\
English $\leftrightarrow$ Spanish & 120 & English $\leftrightarrow$ Spanish $\leftrightarrow$ Portuguese & 13 \\
English $\leftrightarrow$ Telugu & 82 & Spanish $\leftrightarrow$ French $\leftrightarrow$ Portuguese & 8 \\
English $\leftrightarrow$ Tagalog & 73 & Catalan $\leftrightarrow$ Spanish $\leftrightarrow$ Portuguese & 7 \\
English $\leftrightarrow$ Indonesian & 71 & Bengali $\leftrightarrow$ English $\leftrightarrow$ Telugu & 6 \\
Indonesian $\leftrightarrow$ Malay & 65 & Bengali $\leftrightarrow$ Hindi $\leftrightarrow$ Punjabi & 6 \\
\hline
\end{tabular}
\caption{Language combinations with the largest number of shared misinformation claims.}
\label{tab:language_switches}
\end{table}

Among the most common language pairs are Spanish \& Portuguese and Hindi \& Punjabi, indicating that misinformation predominantly circulates among related languages. To corroborate this observation, we employ the Glottolog language catalogue, a resource that maps languages to their respective language families \cite{glottolog}. Mapping the language of each fact-check to its language family we found that while 32.26\% of claim clusters are multilingual and 20.7\% of clusters contain fact-checks from multiple countries, only 8.8\% clusters cross language family boundaries.

\subsection{Research Question 4: Fact-checking delays for multilingual claims}

Fact-checkers require substantial time to fact-check a statement depending on the complexity of the topic. Through interviews with various fact-checking organisations, \citet{graves2016rise} find that the time from a claim first emerging to a fact-check being published can range from a day to over a week depending on the type of claim and the funding and structuring of the fact-checking organisation. In contrast, the temporal dynamics of the spread of misinformaiton have been characterized as ``bursty'', where the majority of interactions with information happens in ``a matter of a few days, if not a few hours'' \citep{shin2018diffusion}.
 
While not all fact-checking organisations provide a timestamp for the first occurrence of a claim 35.8\% of all claims in our dataset have an associated claim-date and publishing-date allowing us to investigate the time-lag between the first occurrence of the claim and the date a fact-check is published.

\begin{figure}[!h]
\centering
\includegraphics[width=\columnwidth]{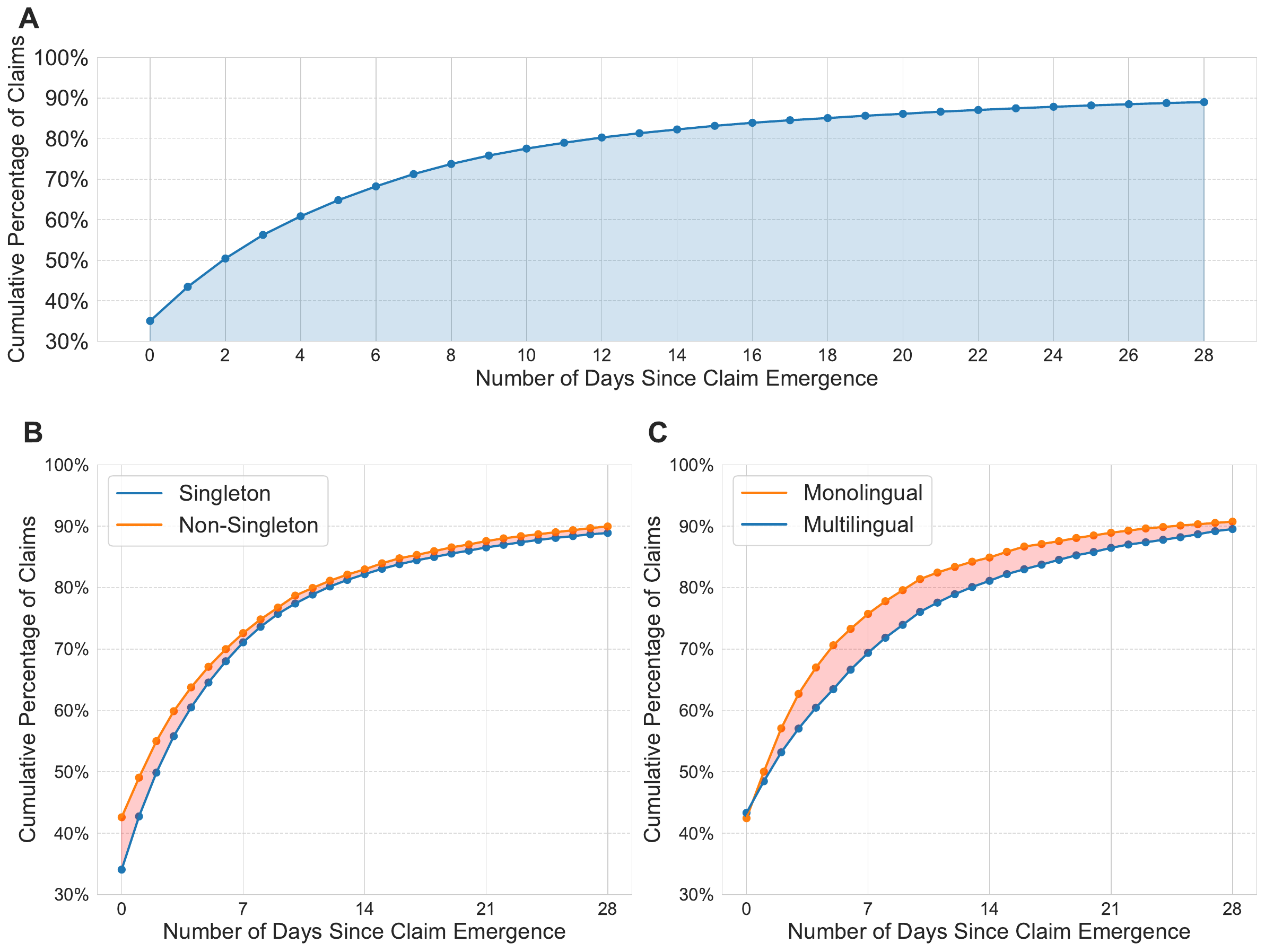}
\caption{\textbf{A} Cumulative percentage of fact-checked claims over time
\textbf{B} Cumulative percentage of clustered fact-checked claims, showing that clustered claims are consistently fact-checked faster than singletons. \textbf{C} Cumulative percentage of clustered fact-checked claims over time by number of languages within cluster. Monolingual clusters are consistently fact-checked faster than multilingual clusters.} 
\label{comparison_cumu}
\end{figure}

Panel A in Figure \ref{comparison_cumu} displays the percentage of claims that have been fact-checked after a given number of days. For example, after 7 days roughly 70\% of claims have been fact-checked. It is important to note that only a minority of fact-checks in our dataset include information on the time that a claim was made. 

In Panel B, the cumulative proportion of fact-checked claims is disaggregated for singletons (claims that have only been fact-checked once) and clustered claims (claims for which we have identified more than one fact-check). There is a small, albeit very consistent, positive difference between the singleton and the non-singleton fact-checks, indicating that the non-singleton claims are generally fact-checked slightly faster than the singletons. 

Panel C disaggregates claims associated with multiple fact-checks into multilingual (claim made in more than one language) and monolingual clusters. Here the difference is similarly consistent but more pronounced. Multilingual fact-checks lag behind monolingual fact-checks over the entire course of the first month. For example, while 73\% of fact-checks corresponding to monolingual claims are published within the first week, multilingual claims need 10 days---or an additional 3 days---to reach the same level. This difference between multilingual and monolingual claims could indicate that fact-checkers are not able to rely on previously published fact-checks in other languages to help them when writing their own assessments.

To check whether the difference in the delays is significant, we conducted both Mann-Whitney U and Log-Rank tests to compare the delays. The one-sided Mann-Whitney U test, which compares central tendencies and tests whether one population has larger values than the other, yielded p-values of $p < 0.001$ for testing if singleton claims have longer delays than clustered claims and $p < 0.001$ for testing if multilingual claims have longer delays than monolingual claims. These results indicate statistically significant differences in the central tendencies of fact-checking delays for both comparisons. As fact-checking can be considered a survival analysis task, where the ``event" is the publication of a fact-check addressing a misinformation claim, we also employed the Log-Rank test, which is specifically designed for time-to-event data and compares entire survival curves. The Log-Rank test resulted in p-values of $p < 0.001$ for singletons vs. clustered claims and $p < 0.001$ for monolingual vs. multilingual claims. The consistency between the two tests strengthens our confidence in these findings, as they offer complementary perspectives on the data: the Mann-Whitney U test focuses on overall distribution differences, while the Log-Rank test captures the temporal aspects of fact-check publication.

\subsection{Research Question 5: Evolution of Misinformation claims}
Understanding the temporal evolution of misinformation claims offers valuable insights into their dynamics. This line of inquiry helps us track how claims change over time, revealing whether they become more or less similar as they spread. By analyzing the factors that influence this evolution, we can start to gain a deeper understanding of the mechanisms that drive the spread of misinformation. This serves as a natural extension to our previous analyses, bridging the gap between static properties and temporal behaviors of misinformation claims.

While we intentionally chose not to restrict clustering based on time---due to the multiple peaks in the temporal diffusion pattern of misinformation, as revealed by \citet{shinDiffusionMisinformationSocial2018}---we find that the majority of edges within each cluster are still closely linked in time, indicating that the corresponding fact-checks were often created near each other temporally.

\begin{figure}[tb]
\centering
\includegraphics[width=\columnwidth]{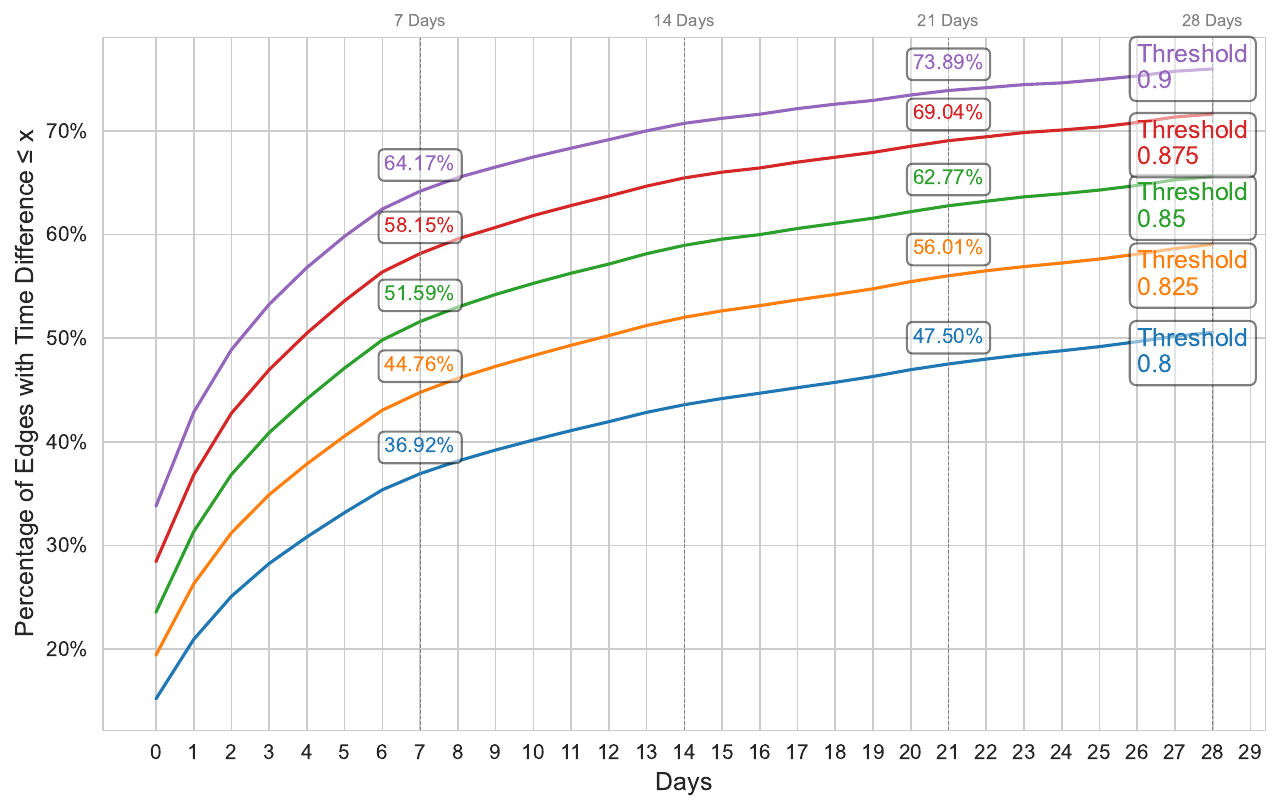}
\caption{Percentage of edges within the same cluster by time difference (in days) between the fact-check dates of the connected claims, for various cosine similarity thresholds used in clustering. For example, at a threshold of 0.875, 58.15\% of edges connect claims with fact-check dates within 7 days of each other, and 69.04\% connect claims with fact-check dates within 21 days.}
\label{EdgesByTime}
\end{figure}

Figure \ref{EdgesByTime} shows the propensity of fact-checks to be closely linked in time. It displays the cumulative percentage of time differences lower than or equal to $x$ days for unconnected nodes within one cluster (i.e., fact-checks checking the same claim). For our chosen cosine similarity threshold of 0.875, 58.15\% of edges have a time difference less than or equal to a week and 69.04\% of edges have a time difference less than or equal to three weeks. Even within each cluster, the time difference of connected nodes is significantly lower than unconnected nodes for all cosine-similarity thresholds.

While directly connected edges always have a cosine-similarity exceeding the pre-set threshold, nodes within the same cluster that are unconnected have lower cosine similarities. We can therefore, inspect how the similarity between unconnected nodes within the same cluster changes over time. A monotonically decreasing cosine similarity would indicate a gradual evolution of the misinformation claim over time. 

\begin{figure}[tb]
\centering
\includegraphics[width=\columnwidth]{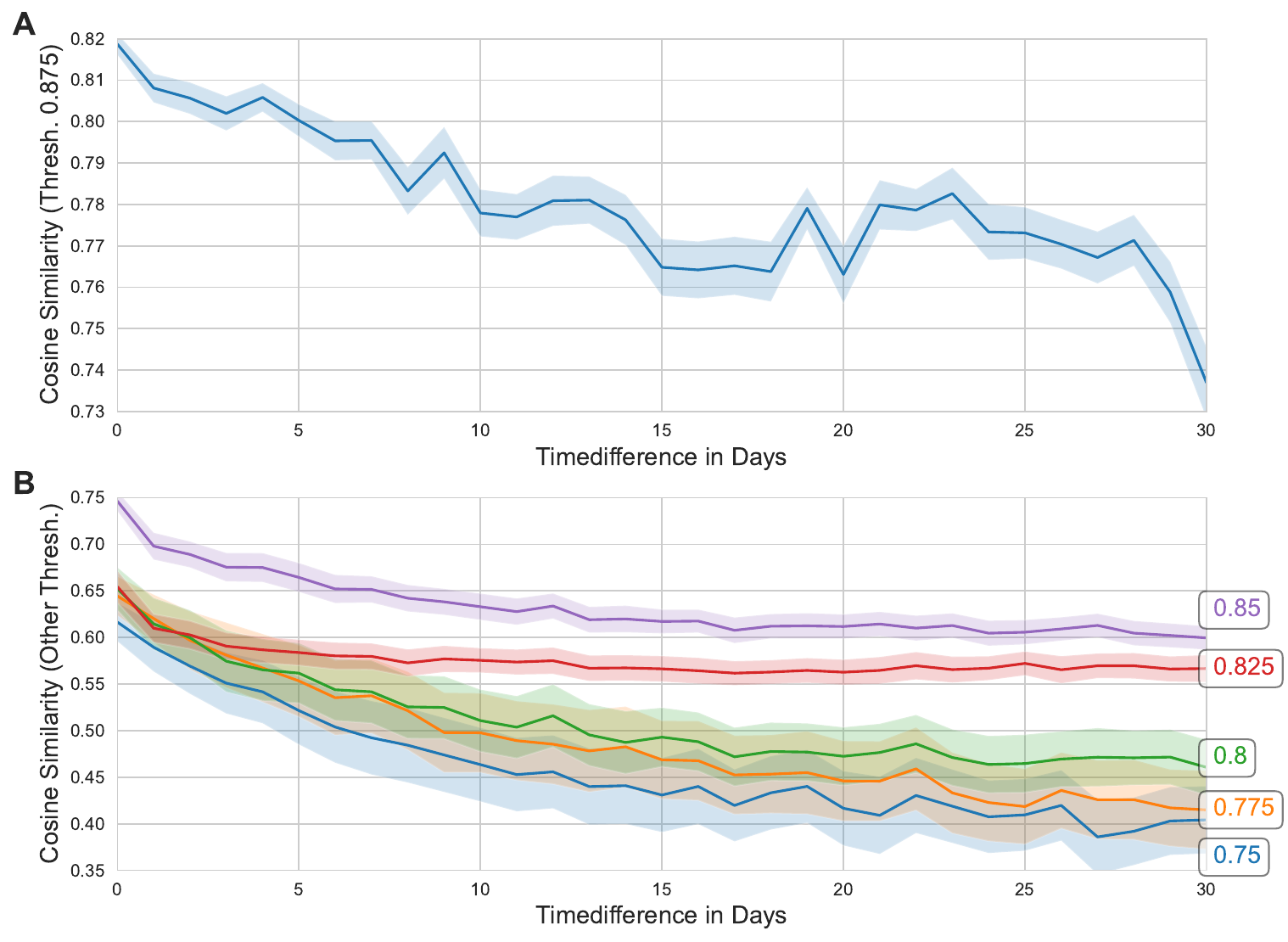}
\caption{Average cosine similarity of unconnected Claims Over Time in the first 30 days. Time measures the number of days between the fact-check dates of two unconnected claims. Shading shows the standard error. \textbf{A} shows the decrease in the cosine similarity for the chosen threshold of 0.875. \textbf{B} Indicates the change in cosine similarity for all lower thresholds.}
\label{fig:cosine-similarity-over-time}
\end{figure}

Panel A of Figure \ref{fig:cosine-similarity-over-time} illustrates the average cosine similarity between all unconnected pairs of nodes within clusters, specifically for our selected cosine-similarity threshold of 0.875. This average is taken across all clusters and plotted as a function of the time difference between the nodes. Panel B in Figure \ref{fig:cosine-similarity-over-time} encompasses all lower cosine-similarity thresholds. Importantly, we only report the average distance for unconnected nodes, as directly connected nodes inherently have a similarity of at least the threshold value. The shading in the Figure represents the standard error. The similarity of unconnected nodes continues to decrease for the first month. For all cases, we see a strong indication of a monotonically decreasing cosine-similarity, suggesting that misinformation claims change over time. A claim changing over time is measured by decreasing cosine similarity between unconnected nodes in the same cluster and signifies gradual semantic drift as the misinformation evolves and is modified during its spread. The same effect is also observed for longer time periods. The effect is consistent for all clustering levels below or equal to our chosen similarity of 0.875. The average similarity between disconnected nodes within the same cluster drops by around 10\% within one year from 0.8 to 0.73 $(t=6.41, p < 0.001)$. 

\begin{figure}[tb]
\centering
\includegraphics[width=\columnwidth]{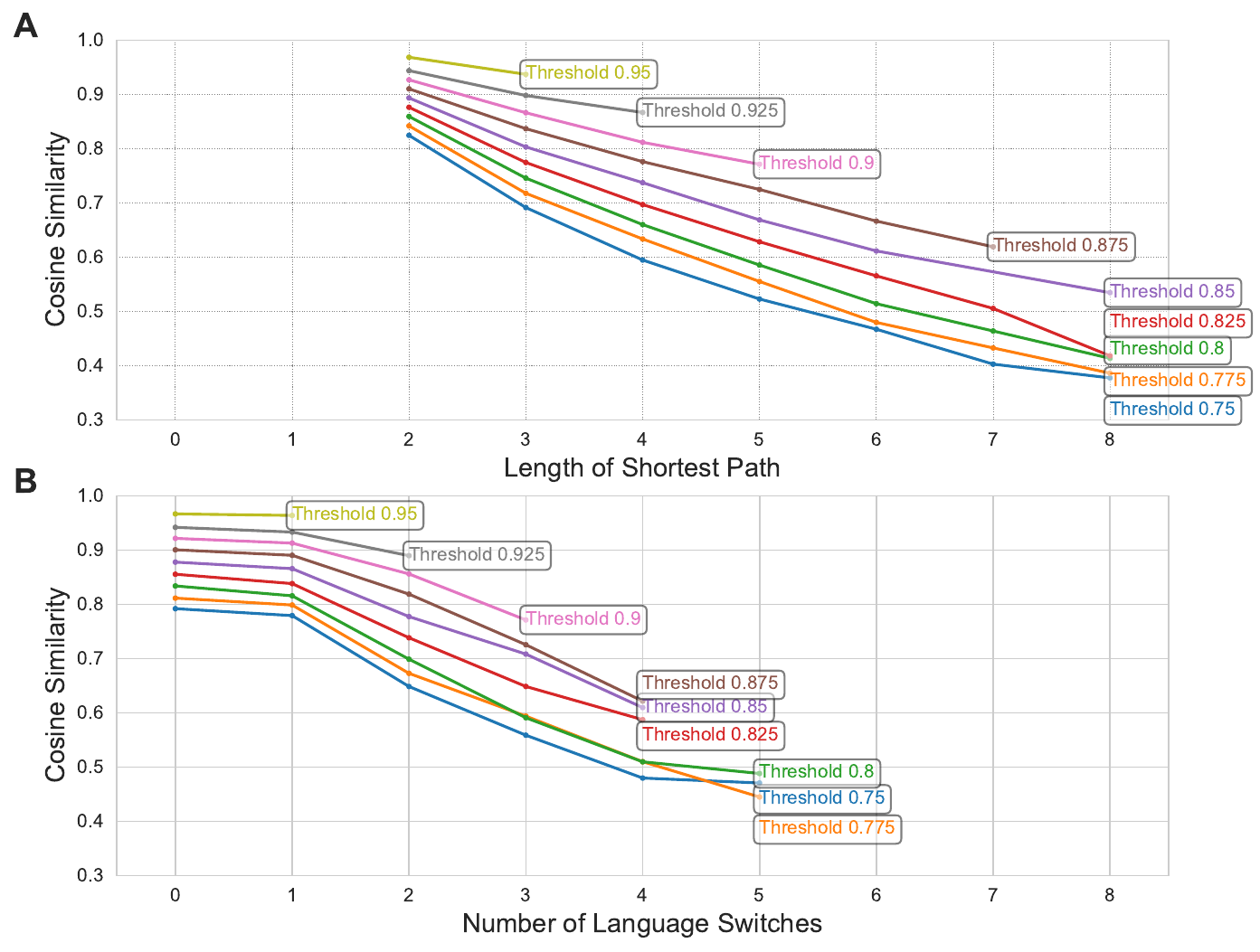}
\caption{Average similarity of most dissimilar Fact-Checks per cluster by length \& number of languages. \textbf{A} shows the decrease in the cosine-similarity of unconnected claims by the length of the shortest path connecting them. \textbf{B} shows the decrease in cosine similarity by the number of language switches on the shortest path.}
\label{AVGSIMIBYLEN}
\end{figure}

Similarly, we can inspect what factors influence the evolution of misinformation claims within clusters. To do that, we determine the two most dissimilar nodes per cluster using the cosine similarity of the embeddings. Subsequently, we determine the shortest path connecting these dissimilar nodes. Figure \ref{exampleCluster} shows an example multilingual cluster. The color of each node refers to the language of the fact-check. The two most dissimilar nodes, F \& A, are circled in black. The shortest path between these two nodes is highlighted in yellow. This way of looking at each cluster allows us to investigate how misinformation claims within the same cluster evolve. Figure \ref{AVGSIMIBYLEN} shows the average cosine similarity of the two most dissimilar claims in each cluster plotted against the number of language switches and the shortest path length. We only plot observations with at least 100 valid pairs. 
 
We find that the average cosine similarity of the two most dissimilar nodes varies considerably with the length of the shortest path (i.e., tracing the most probable path of evolution). Both the number of unique languages and the number of language switches are highly significant predictors of changes in the cosine similarity. 

\section{Limitations}

Fact-checks are not an unbiased lens into misinformation worldwide. Fact-checkers necessarily exert subjectivity in deciding what claims to investigate and tend to focus on high-profile viral claims that gain widespread traction or more consequential claims. This analysis only considers claims that have been fact-checked, which is not representative of all claims circulating online. The methodology used may not group all fact-checks of the same claim together. While manual analysis finds nearly all clusters are coherent and only consist of one specific claim, it cannot be proven that all fact-checks checking one claim necessarily are in one cluster. We cluster claims as reported by fact-checkers, but the actual text of the misinformation can be edited by fact-checkers who may interpret or contextualize them. The rephrasing of claims may induce biases into the data. 
Fact-checks have several additional limitations that constrain their ability to fully capture misinformation dynamics. 
Fact-checking organizations can be influenced by varying missions, funding incentives, and the different levels of attention required for different types of claims. 
Additionally, the availability of fact-checks varies by region based on where fact-checking initiatives exist. However, despite these limitations, fact-checks remain a useful proxy for studying global misinformation by documenting the details of specific dubious claims. They offer insights into the spread, subject matter, and timeline of misinformation. The diverse dataset of 264,487 unique fact-checks in 95 languages underscores the global and multilingual endeavours of fact-checking, testifying to the universal scope of the problem of inaccurate information across geographical, linguistic, and cultural boundaries. In contrast to focusing on a single social media site, which may overlook misinformation in countries where that platform is not dominant, the study of fact-checks transcends these boundaries. The number of fact-checks and the worldwide distribution of fact-checking organizations further solidify the case for using fact-checks as a proxy to study the undercurrents of misinformation across various social media platforms. The analysis of the time differences between claim emergence and fact-check publishing only considers a minority of fact-checks, as only 36\% in the dataset include information on the time that a claim was made, potentially introducing a selection bias. Additionally, the comparison of multilingual and monolingual claims focuses on the minority of the data that is clustered, which may not be generalizable to all claims. The reasons behind a fact-check having or not having a claim date are unknown, and this difference might be influenced by factors such as the ease of fact-checking, the importance of the claim, or its source.

\section{Discussion \& Conclusion} 
This paper investigates fact-checking in the online multilingual space. Above all, we find that while most misinformation claims are only fact-checked once, 10.26\% of misinformation claims are fact-checked multiple times. This observation highlights the existence of a subset of claims that undergo scrutiny by multiple organizations. It suggests a persistent pattern where certain claims, due to their nature, importance, or controversy, command attention from multiple fact-checking organizations. Our token analysis further illuminates this phenomenon. We find some words that are more likely to be appear in claims in multiple languages and in claims that are fact-check multiple times. Such words include ``Pfizer,'' ``Ivermectin,'' and words related to conspiracy theories. These topics present an opportunity for more efficient allocation of fact-checking resources to minimize repeated work.

Next, we find that 32.26\% of misinformation claims that are fact-checked more than once are checked in multiple languages. Nevertheless, misinformation still diffuses predominantly within the same language. Even claims that do travel across languages tend to do so in the same country or in a closely related language. This echoes the culture proximity theory that culture and language are still very influential factors increasing online users' information consumption. Though technologies such as machine translation and social media make it easier for cross-language communication, they do not necessarily function that way in everyday use. In other words, our research highlights the importance of local fact-checking. Nonetheless, a notable amount of misinformation spreads across languages, which simultaneously highlights the importance of global cooperation. 

The analysis of fact-checking delays motivates our study and highlights the need for more technical tools and experts to evaluate multinational and multilingual misinformation. The consistent differences in fact-checking speed between monolingual and multilingual claims, as well as between clustered and unclustered claims, suggest that there is room for improvement in the fact-checking process by developing tools for fact-checkers to quickly assess whether a claim has previously been addressed in another language. There are two possible explanations for the delay between clustered and unclustered claims. First, claims which are fact-checked more than once could generally be more time sensitive, more important or easier to verify. Second, fact-checkers may be leveraging work done by other fact-checking organizations when writing their assessment leading to a reduced workload and therefore faster publishing times. By developing tools and expertise specifically designed to address the challenges posed by multilingual and cross-border misinformation, we can work towards more efficient and effective fact-checking practices. This, in turn, can help combat the spread of false information and promote a more informed global discourse.

Moreover, we show that misinformation claims can change over time and changes are especially common when a claim is found in multiple languages. Our data, however, only contains the misinformation claims as reported by fact-checkers. There is some degree of editorial voice in how fact-checkers write or phrase misinformation claims; so, future work should seek to analyze a global dataset of the original misinformation posts on social media. Such posts are often removed, which makes this a challenge and will require closer collaboration between academics and fact-checkers. While our results show claims change over time, the specific mechanisms and consequences remain unclear. Future research should delve more deeply into how claims change and the role of cross-lingual spread in this process, which may help fact-checkers better anticipate how claims will change and re-occur.

Based on our findings, we can make several practical recommendations for fact-checking generally. First, since much misinformation is local, we need local fact-checking initiatives and cannot rely on organizations in other countries to fact-check content even if they share a common language.

Second, given the notable amount of cross-lingual misinformation, fact-checking organizations could prioritize developing systems to check if a claim has already been fact-checked in any language before beginning work. We note, for instance, that the web interface to Google's data, Google Fact Check Tools - Explorer, does not return results across languages. That is, it does not perform cross-lingual information retrieval currently, although the methods used in this paper demonstrate this is possible. Our analysis suggests such systems could help reduce the additional delays we observed for multilingual claims.

Finally, our analysis of language pairs (e.g., Spanish--Portuguese, Hindi--Punjabi) suggests natural partnership opportunities between fact-checking organizations working in related languages, which could facilitate faster and more efficient fact-checking across linguistic boundaries.

\section*{Declarations}
\subsection*{Availability of data and material}
All data-preparation and analysis code are uploaded to \url{https://github.com/dorianquelle/Lost-In-Translation}. The dataset used in the analysis is available under \url{https://zenodo.org/records/11098780}. The dataset consists of the link to each fact-check, and the membership of each fact-check to any clusters at all cosine similarity threshold levels. In addition, we have uploaded an .npy file linking each claim to its embedding.

\subsection*{Competing interests}
SAH consults for Meedan, a non-profit organization that provides software and programmatic activities for fact-checking organizations.

\subsection*{Funding}
This research is supported by a grant from The Alan Turing Institute for the project entitled ``Effective discovery, tracking, and response to mis- and disinformation.''

\subsection*{Authors' contributions}
SAH and DQ  conceived of the research question and structure of the  manuscript. DQ performed all computations. CC contributed theoretical and qualitative analyses and surveyed the literature. AB provided methodological insights. AB and SAH supervised the findings of this work. All authors discussed the results and contributed to the final manuscript.

\subsection*{Acknowledgments}
We would like to thank Devin Gaffney for building and maintaining the Fetch service from which our crawled fact-check data came.


\newpage

\section*{Abbreviations}
\begin{itemize}
\item ANNOY - Approximate Nearest Neighbors Oh Yeah!
\item API - Application Programming Interface
\item CSS - Cascading Style Sheets
\item DBSCAN - Density-Based Spatial Clustering of Applications with Noise
\item HDBScan - Hierarchical Density-Based Spatial Clustering of Applications with Noise
\item HTML - HyperText Markup Language
\item IFCN - International Fact-Checking Network
\item LaBSE - Language-agnostic BERT Sentence Embedder
\item LSH - Locality Sensitive Hashing
\item MLM - Masked Language Modeling
\item NLTK - Natural Language Toolkit
\item NLP - Natural Language Processing
\item TLM - Translation Language Modeling
\item UMAP - Uniform Manifold Approximation and Projection
\end{itemize}

\section{Appendix A} \label{Appendix:A}
Two major events which have the potential to significantly alter the information environment on social media fall into our observation period. First, the outbreak of Covid-19 and secondly the invasion of Ukraine by Russia. We examined the impact of COVID-19 and the Ukraine war on our results using case-insensitive keywords. To identify fact-checks related to Russias invasion of Ukraine we filtered for fact-checks containing the keywords \emph{kiev, ukraine, zelensky, kyiv, russia, mocow,} and \emph{putin}. To identify fact-checks related to Covid-19 we used the terms \emph{covid, vaccine, vaccination, corona}. The keywords were matched against the machine translations of all claims to English.

\begin{figure}[!h]
    \centering
    \includegraphics[width=0.9\linewidth]{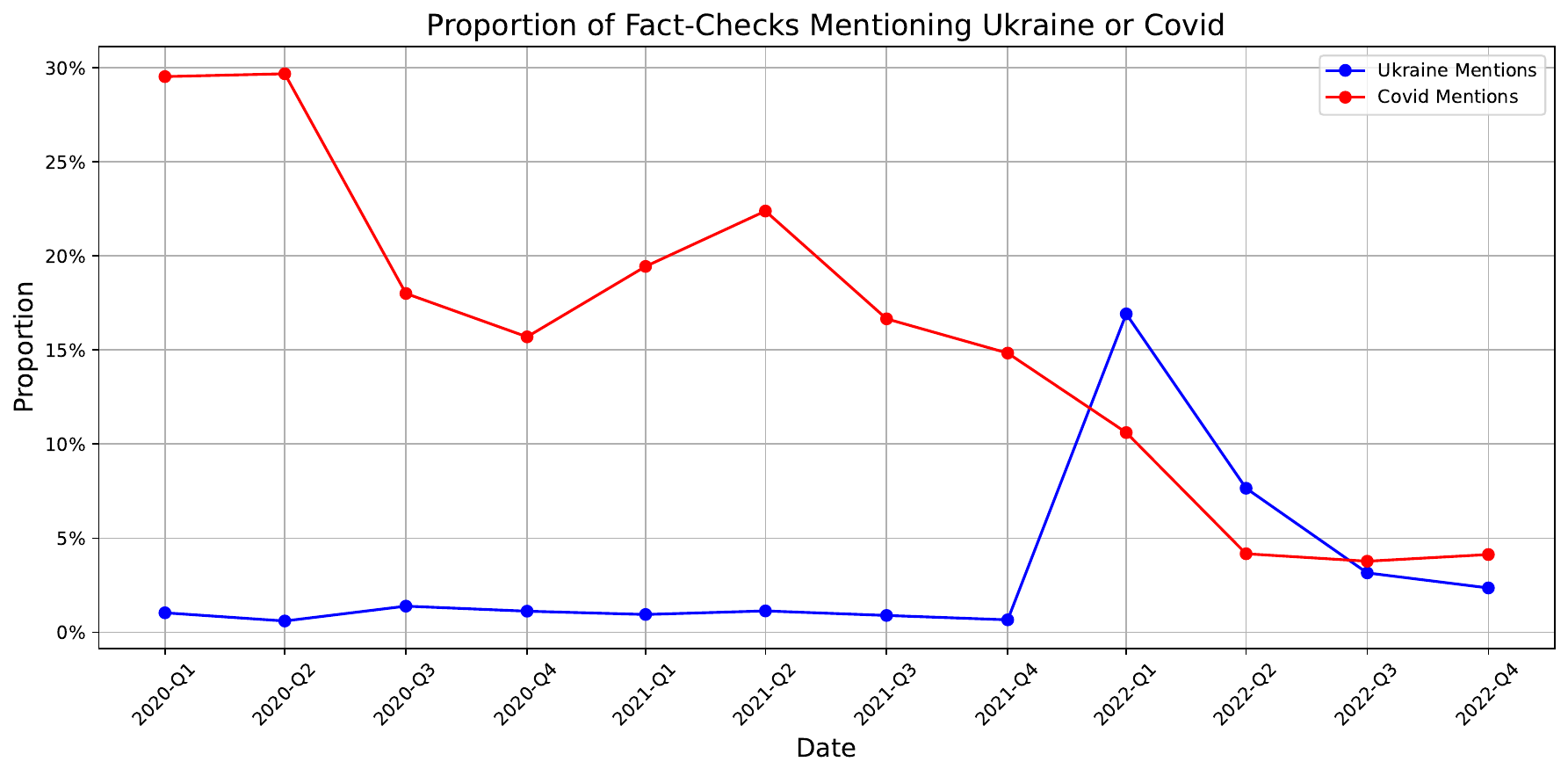}
    \caption{Proportion of Fact-Checks mentioning keywords related to Covid-19 (\emph{covid, vaccine, vaccination, corona}) and the invasion of Ukraine (\emph{kiev, ukraine, zelensky, kyiv, russia, mocow,} and \emph{putin}). During the first half of 2020 Covid-19 accounts for nearly 30\% of the fact-checks in our dataset but declines to less than 5\% by the end of the time period we study. Conversely, there is little coverage related to Ukraine until Russia started its invasion in Q1 of 2022. The proportion of fact-checks mentioning Ukraine is higher than that mentioning Covid-19 at the start of the conflict before rapidly declining.}
    \label{fig:covid_ukraine_mentions}
\end{figure}

During the first half of 2020, Covid-19 related keywords were present in 29\% of fact-checks in our dataset. This proportion then began to steadily decline. By the end of our observation period, Covid-19 related fact-checks accounted for less than 5\% of the data. In contrast, Ukraine-related fact-checks were non-existent until Russia initiated its invasion in Q1 2022. At this point, the proportion of fact-checks mentioning Ukraine briefly surpassed Covid-19, reaching 17\%. However, this coverage rapidly diminished, eventually stabilizing at 2\% of all fact-checks.
Both topics were less likely to be singletons, as revealed by independent samples t-tests $(\text{COVID-19}: \text{t} = -23.77, \text{p} < 0.001; \text{Ukraine}: \text{t} = -6.73, \text{p} < 0.001)$, showing that misinformation related to these topics was more likely to be fact-checked more than once. Non-singleton COVID-19 \& Ukraine claims were significantly more likely to be multilingual, again demonstrated by t-tests $(\text{COVID-19}: \text{t} = 20.70, \text{p} < 0.001; \text{Ukraine}: \text{t} = 8.34, \text{p} < 0.001)$. The exclusion of both all claims related to Ukraine and all claims related to Covid-19 reduces the percentage of non-clustered claims from 10.26\% to 9.46\% and the percentage of clustered claims that are multilingual from 32.26\% to 30.63\%.

\end{document}